\documentclass{article}

\usepackage{graphicx}
\usepackage{amsmath}
\usepackage{amssymb}
\usepackage{booktabs}

\usepackage{paralist}

\usepackage{makecell}
\usepackage{subcaption}
\usepackage{multirow}
\usepackage{pifont}

\usepackage{bbm}
\usepackage{mathtools}
\usepackage[table]{xcolor}

\usepackage[preprint]{corl_2022} 

\title{VectorFlow: Combining Images and Vectors for Traffic Occupancy and Flow Prediction}

\author{
  Xin Huang$^{1}$,\; Xiaoyu Tian$^{2}$,\; Junru Gu$^{2}$,\; Qiao Sun$^{2}$,\; Hang Zhao$^{2}$\\
  $^{1}$CSAIL, MIT \quad $^{2}$IIIS, Tsinghua University\\
  \texttt{xhuang@csail.mit.edu},\quad \texttt{hangzhao@mail.tsinghua.edu.cn} \\
}

\begin{document}
\maketitle


\begin{abstract}
    Predicting future behaviors of road agents is a key task in autonomous driving. While existing models have demonstrated great success in predicting marginal agent future behaviors, it remains a challenge to efficiently predict consistent joint behaviors of multiple agents. Recently, the occupancy flow fields representation was proposed to represent joint future states of road agents through a combination of occupancy grid and flow, which supports efficient and consistent joint predictions. In this work, we propose a novel occupancy flow fields predictor to produce accurate occupancy and flow predictions, by combining the power of an image encoder that learns features from a rasterized traffic image and a vector encoder that captures information of continuous agent trajectories and map states. The two encoded features are fused by multiple attention modules before generating final predictions. Our simple but effective model ranks 3rd place on the Waymo Open Dataset Occupancy and Flow Prediction Challenge, and achieves the best performance in the occluded occupancy and flow prediction task.
\end{abstract}



\section{Introduction}
Behavior prediction is a key task in autonomous driving, as it enables autonomous vehicles to interact with other road agents in crowded scenes. Many existing work focuses on predicting accurate marginal agent trajectories~\cite{gao2020vectornet,gu2021densetnt,varadarajan2022multipath++,huang2022hyper}, yet they do not account for agent interactions in the future and could lead to inconsistent predictions~\cite{sun2022m2i}. On the other hand, predicting joint trajectories of multiple agents remains an open challenge, as the number of future trajectory combinations grows exponentially with the number of agents. Despite recent progress toward joint trajectory prediction~\cite{sun2022m2i,ngiam2021scene}, it remains an open question on predicting joint behaviors in an efficient and scalable way to meet the latency requirements for real-time deployment in crowded urban scenes~\cite{kim2022stopnet}.

Recently, the occupancy flow fields representation~\cite{mahjourian2022occupancy} was introduced to overcome the shortcomings of the traditional trajectory representation for behavior prediction of multiple agents. This new representation is a spatio-temporal grid, where each grid cell includes i) the probability of the cell occupied by any agent and ii) the flow representing the motion of the agents that occupy the cell. It offers better efficiency and scalability, as the computational complexity of predicting occupancy flow fields is independent of the number of road agents in the scene~\cite{kim2022stopnet}.

In this work, we propose a novel occupancy flow fields predictor by exploring the benefit of combining vectorized and rasterized representations as the input to the predictor. Both representations have demonstrated great success in trajectory prediction benchmarks, yet few models combine them together. Our proposed model, \emph{VectorFlow}, is a simple but effective approach that fuses both vectorized and rasterized representations of traffic context through attention to predict occupancy flow fields for both observed agents and occluded agents. 

Experimental results show that VectorFlow achieves state-of-the-art performance in the occluded occupancy and flow prediction task, and ranks 3rd place on the Waymo Open Dataset Occupancy and Flow Prediction Challenge.


\section{Problem Formulation}
\label{sec:problem}

In this work, we use the following problem setup\footnote{More detail on the data format can be found at: \url{https://waymo.com/open/challenges/2022/occupancy-flow-prediction-challenge/}}: Given one-second history of traffic agents in a scene and the scene context such as map coordinates~\cite{ettinger2021large}, the objective is to predict i) future observed occupancy, ii) future occluded occupancy, and iii) future flow of all the vehicles in a scene over the next 8 waypoints, where each waypoint covers a one-second interval.

\subsection{Data Processing}
We process the input into a rasterized image and a set of vectors. To obtain the image, we create a rasterized grid at each time step in the past given the observed agent trajectories and map data, with respect to the local coordinate of the self-driving car (SDC)\footnote{More detail can be found at \url{https://github.com/waymo-research/waymo-open-dataset/blob/master/tutorial/tutorial_occupancy_flow.ipynb}}. To obtain the vectorized input that is consistent with the rasterized image, we follow the same transformation by rotating and shifting the input agent and map coordinates with respect to SDC's local view.

\vspace{2mm}
\section{Model}
\label{sec:model}
In this section, we provide a brief overview of VectorFlow, which adopts a standard encoder-decoder architecture. More implementation detail can be found in Sec.~\ref{sec:result}.

\subsection{Encoder}
The encoder includes two parts: a VGG-16 model~\cite{simonyan2014very} that encodes the rasterized representation, and a VectorNet~\cite{gao2020vectornet} model that encodes the vectorized representation. We fuse the vectorized features with the features of the last two stages of VGG-16 by cross-attention modules. The fused features are upsampled to the original resolution as the input rasterized features by an FPN-style network. More detail can be found in Fig.~\ref{fig:approach} and Sec.~\ref{sec:detail}.

\subsection{Decoder}
The decoder is a single 2D convolution layer that maps the output of the encoder to the occupancy flow fields prediction, which includes a sequence of 8 grid maps representing the predicted occupancy and flow at each time step over the next 8 seconds.

\subsection{Loss}
We follow the loss function as in~\cite{mahjourian2022occupancy}, including a cross entropy loss on observed occupancy prediction $\mathcal{L}_{O,b}$, a cross entropy loss on occluded occupancy prediction $\mathcal{L}_{O,c}$, and an L2 loss on flow prediction $\mathcal{L}_{F}$. The total loss is:
\begin{equation}
    \mathcal{L} = \alpha \mathcal{L}_{O,b} + \beta \mathcal{L}_{O,c} + \gamma \mathcal{L}_{F}.
\end{equation}
	
\vspace{2mm}
\section{Experiment}
\label{sec:result}

\subsection{Dataset}
We train, evaluate, and test our model on the Waymo Open Motion Dataset (WOMD) based on the standard split and the filtered scenarios.

\subsection{Metrics}
We follow the metrics proposed by~\cite{mahjourian2022occupancy}, which include \emph{AUC} and \emph{Soft IoU} for observed occupancy, occluded occupancy, and flow-grounded occupancy, as well as the \emph{end-point error (EPE)} that measures the error of flow prediction.

\subsection{Model Detail}
\label{sec:detail}
\begin{figure}[t]
    \centering
    \includegraphics[width=1\textwidth]{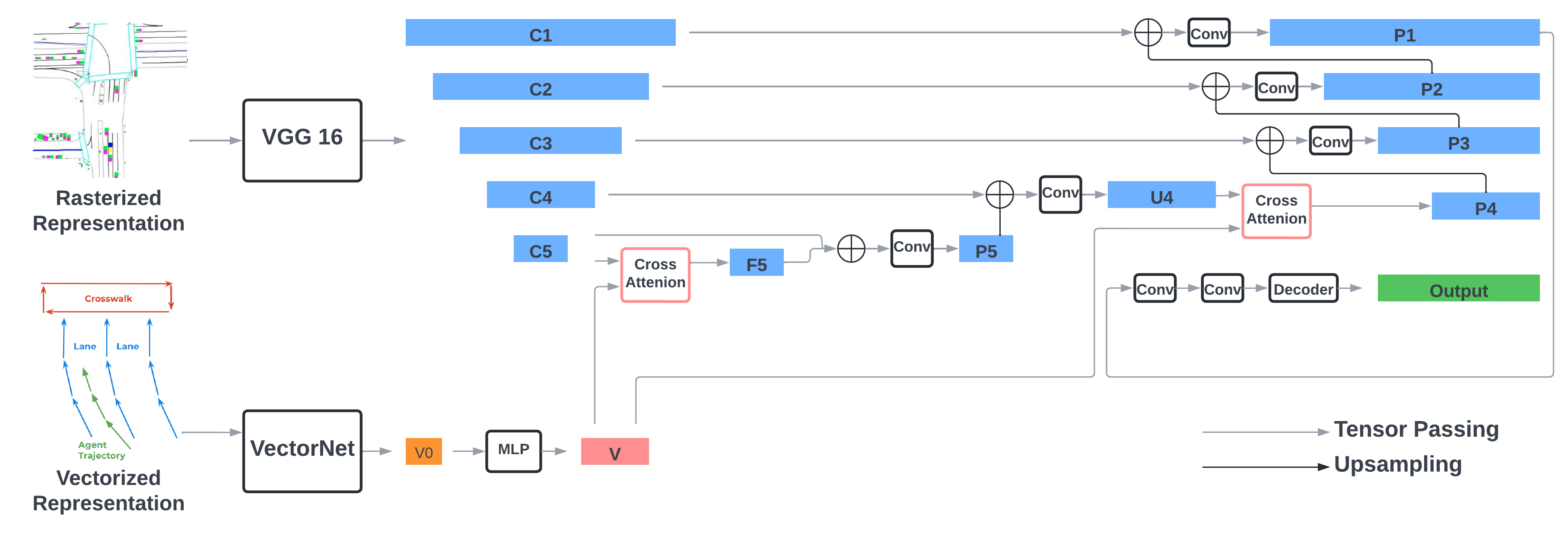}
    \caption{Overview of VectorFlow. More detail can be found in~Sec.~\ref{sec:detail}. Representation images credit to~\cite{gao2020vectornet}.}
    \label{fig:approach}
\end{figure}

Our model is illustrated in Fig~\ref{fig:approach}.
We use the standard VGG-16 model from \emph{torchvision.models} as our rasterized encoder, and follow the implementations of VectorNet as in~\cite{gu2021densetnt}\footnote{Code available at \url{https://github.com/Tsinghua-MARS-Lab/DenseTNT}}. 
The input to the VectorNet includes i) a set of road element vectors with a shape of $B \times N_R \times 9$, where $B$ is the batch size, $N_R=10000$ is the maximum number of road element vectors, and the last dimension of 9 represents the positions ($x, y$) and headings ($\cos\theta, \sin\theta)$ of two end points in each vector and the vector id; ii) a set of agent vectors with a shape of $B \times 1280 \times 9$, including the vectors of up to 128 agents in a scene, where each agent has 10 vectors from the observed positions. We follow VectorNet by first running a local graph over each traffic element based on their ids and second running a global graph over all local features to obtain a vectorized feature with a shape of $B \times 128 \times N$, where $N$ is the total number of traffic elements, including road elements and agents. We further quadrupled the size of the feature through an MLP layer to obtain a final vectorized feature $V$ with a shape of $B \times 512 \times N$, so that its feature size is consistent with the channel size of the image feature, as discussed in the following paragraph.

We denote the output features of each VGG stage as $\{C_1,C_2,C_3,C_4,C_5\}$, and they have strides of $\{1,2,4,8,16\}$ pixels with respect to the input image and a hidden dimension of 512. The vectorized feature $V$ is fused with the rasterized image feature $C_5$ with a shape of $B \times 512 \times 16 \times 16$ by a cross-attention module to obtain $F_5$ with the same shape. The query term of the cross-attention is the image feature $C_5$ flattened into $B \times 512 \times 256$ with 256 tokens, and the key and value term is the vectorized feature $V$ with $N$ tokens.
We then concatenate $F_5$ and $C_5$ in the channel dimension and pass it through two $3 \times 3$ conv layers to obtain $P_5$ with a shape of $B \times 512 \times 16 \times 16$. $P_5$ is upsampled and concatenated with $C_4$ ($B \times 512 \times 32 \times 32$) by an FPN-style $2 \times 2$ upsampling module to generate $U_4$ with the same shape as $C_4$. 
Next, we perform another round of fusion between $V$ and $U_4$ to obtain $P_4$ ($B \times 512 \times 32 \times 32$) following the same procedure, including cross attention.
At the end, $P_4$ will be gradually upsampled by the FPN-style network and concatenated with $\{C_3,C_2,C_1\}$ to generate $P_1$ with a shape of $B \times 512 \times 256 \times 256$. We pass $P_1$ through two $3 \times 3$ conv layers to obtain the final output feature with a shape of $B \times 128 \times 256 \times 256$.

The decoder is a single 2D convolution layer with an input channel size of 128 and an output channel size of 32 (8 waypoints $\times$ 4 output dimensions).

\begin{table*}[t!]
    \centering
    \footnotesize
    \bgroup
    \def\arraystretch{1.05}%
    \begin{tabular}{lcc|cc|cc|c}
    \toprule
    & \multicolumn{2}{c|}{Observed} & \multicolumn{2}{c|}{Occluded} & \multicolumn{2}{c|}{Flow-Grounded} & \\
    Model & AUC $\uparrow$ & Soft IoU $\uparrow$ & AUC $\uparrow$ & Soft IoU $\uparrow$ & \cellcolor{gray!50}AUC $\uparrow$ & Soft IoU $\uparrow$ & EPE $\downarrow$ \\
    \midrule
    HorizonOccFlowPred. & \textbf{0.803} & 0.235 & 0.165 & 0.017 & \textbf{0.839} & \textbf{0.633}  & 3.672 \\
    Look Around & 0.801 & 0.234 & 0.139 & 0.029 & 0.825 & 0.549  & \textbf{2.619} \\
    Temporal Query & 0.757 & 0.393 & 0.171 & 0.040 & 0.778 & 0.465  & 3.308 \\
    STrajNet & 0.751 & 0.482 & 0.161 & 0.018 & 0.777 & 0.555  & 3.587 \\
    3D-STCNN & 0.691 & 0.412 & 0.115 & 0.021 & 0.733 & 0.468  & 4.181 \\
    Motionnet & 0.694 & 0.411 & 0.141 & 0.031 & 0.732 & 0.469  & 4.275 \\
    FTLS & 0.618 & 0.318 & 0.085 & 0.019 & 0.689 & 0.431  & 9.612 \\
    OccFlowNet & 0.667 & 0.391 & 0.111 & 0.026 & 0.678 & 0.443  & 6.636 \\
    \midrule
    VectorFlow & 0.755 & \textbf{0.488} & \textbf{0.174} & \textbf{0.045} & 0.767 & 0.530  & 3.583 \\
    \bottomrule
    \end{tabular}%
    \egroup
    \caption{Prediction performance on the test set. The best performed metrics are bolded and the grey cell indicates the ranking metric used by the WOMD benchmark. Our model achieves the best performance in three metrics.}
    \label{tab:results}
\end{table*}

\begin{table*}[t!]
\vspace{2mm}
    \centering
    \footnotesize
    \bgroup
    \def\arraystretch{1.05}%
    \begin{tabular}{lcc|cc|cc|c}
    \toprule
    & \multicolumn{2}{c|}{Observed} & \multicolumn{2}{c|}{Occluded} & \multicolumn{2}{c|}{Flow-Grounded} & \\
    Model & AUC $\uparrow$ & Soft IoU $\uparrow$ & AUC $\uparrow$ & Soft IoU $\uparrow$ & \cellcolor{gray!50}AUC $\uparrow$ & Soft IoU $\uparrow$ & EPE $\downarrow$ \\
    \midrule
    VectorFlow (VGG-only) & 0.746 & 0.468 & 0.139 & 0.034 & 0.755 & 0.520  & 3.713 \\
    VectorFlow & \textbf{0.760} & \textbf{0.490} & \textbf{0.173} & \textbf{0.050} & \textbf{0.761} & \textbf{0.524}  & \textbf{3.603} \\
    \bottomrule
    \end{tabular}%
    \egroup
    \caption{Prediction performance on the validation set. The best performed metrics are bolded. Using a vectorized representation helps boost the prediction performance.}
    \label{tab:ablation}
\end{table*}


\subsection{Training Detail}
We train our model on the full training set of WOMD with a batch size of 32 for 16 epochs on 8 Nvidia A10 GPUs. 
We use an Adam optimizer and a learning rate scheduler that decays the learning rate by 50\% every 5 epochs, with an initial value of 1e-3. The loss coefficients are $\alpha = \beta = 1000$, and $\gamma = 1$, as customary in~\cite{mahjourian2022occupancy}.

\subsection{Results}
We present the results of our model and other entries in the Waymo Challenge in Table~\ref{tab:results}. Our model achieves the best performance in three metrics, including AUC and Soft IoU scores for occluded occupancy predictions, and Soft IoU score for observed occupancy predictions. The gap in AUC scores is partially due to the choice of our loss function, compared to other entries that use a focal loss. Our model ranks 3rd place on the Waymo Open Dataset Occupancy and Flow Prediction Challenge. 

Furthermore, we compare the performance of our model with a variant that only includes the VGG encoder. The results in Table~\ref{tab:ablation} show that fusion helps improve the prediction performance, especially in occluded metrics. More specifically, the Occluded AUC and Occluded Soft IoU scores improve by 24.46\% and 47.06\%, respectively.


\section{Conclusion}
In this work, we present a simple but effective occupancy and flow predictor that efficiently generates joint agent behaviors as occupancy flow fields. Our predictor VectorFlow fuses two representations commonly used in trajectory prediction, vectorized representation and rasterized representation, through multiple attention modules. It achieves the state-of-the-art performance on the Waymo Open Dataset Occupancy and Flow Prediction Challenge. 
\label{sec:conclusion}




\bibliography{report}  

\end{document}